\documentclass[sigconf]{acmart}

\pdfoutput=1

 \copyrightyear{2020}
 \acmYear{2020}
 \setcopyright{rightsretained}
 \acmConference[RecSys '20]{Fourteenth ACM Conference on Recommender Systems}{September 22--26, 2020}{Virtual Event, Brazil}
\acmDOI{10.1145/3383313.3412215}
\acmISBN{978-1-4503-7583-2/20/09}

\usepackage{float}
\usepackage{soul}
\usepackage{color}
\usepackage{amsmath}
\usepackage{bbm}

\newcommand{\specialcell}[2][c]{%
  \begin{tabular}[#1]{@{}c@{}}#2\end{tabular}}
 
\definecolor{Blue}{rgb}{0,0,1}

\definecolor{Orange}{rgb}{1,0.5,0}

\definecolor{Green}{rgb}{0,1,0}

\DeclareMathOperator{\sign}{sign}

\begin{document}

\title[Free Lunch! Retrospective Uplift Modeling]{Free Lunch! Retrospective Uplift Modeling for Dynamic Promotions Recommendation within ROI Constraints}

\author{Dmitri Goldenberg}
\authornote{Both authors contributed equally to this research.}
\email{dima.goldenberg@booking.com}
\author{Javier Albert}
\authornotemark[1]
\email{javier.albert@booking.com}
\affiliation{\institution{Booking.com, Tel Aviv, Israel}}
\author{Lucas Bernardi}
\authornote{Corresponding author.}
% \affiliation{\institution{Booking.com, Amsterdam, Netherlands}}
\email{lucas.bernardi@booking.com}
\author{Pablo Estevez}
\email{pablo.estevez@booking.com}
\affiliation{\institution{Booking.com, Amsterdam, Netherlands}}

\begin{abstract}
%% revising paper checklist:
%% https://web.cs.dal.ca/~eem/gradResources/KDD/Checklist%20for%20Revising%20a%20SIGKDD%20Data%20Mining%20Paper.pdf

Promotions and discounts have become key components of modern e-commerce platforms. For online travel platforms (OTPs), popular promotions include room upgrades, free meals and transportation services. By offering these promotions, customers can get more value for their money, while both the OTP and its travel partners may grow their loyal customer base. However, the promotions usually incur a cost that, if uncontrolled, can become unsustainable. Consequently, for a promotion to be viable, its associated costs must be balanced by incremental revenue within set financial constraints. Personalized treatment assignment can be used to satisfy such constraints.
This paper introduces a novel uplift modeling technique, relying on the Knapsack Problem formulation, that dynamically optimizes the incremental treatment outcome subject to the required Return on Investment (ROI) constraints. The technique leverages \textit{Retrospective Estimation}, a modeling approach that relies solely on data from positive outcome examples. The method also addresses training data bias, long term effects, and seasonality challenges via online-dynamic calibration. This approach was tested via offline experiments and online randomized controlled trials at Booking.com - a leading OTP with millions of customers worldwide, resulting in a significant increase in the target outcome while staying within the required financial constraints and outperforming other approaches.

\end{abstract}

\begin{CCSXML}
<ccs2012>
<concept>
<concept_id>10010147.10010257.10010282.10010292</concept_id>
<concept_desc>Computing methodologies~Learning from implicit feedback</concept_desc>
<concept_significance>500</concept_significance>
</concept>
<concept>
<concept_id>10002951.10003260.10003261.10003271</concept_id>
<concept_desc>Information systems~Personalization</concept_desc>
<concept_significance>500</concept_significance>
</concept>
</ccs2012>
\end{CCSXML}

\ccsdesc[500]{Computing methodologies~Learning from implicit feedback}
\ccsdesc[500]{Information systems~Personalization}

\keywords{Uplift Modeling, Machine Learning, Causal Inference, Benefits Recommendation, Implicit Feedback}

\maketitle

\section{Introduction}\label{sec:intro}
Modern online travel platforms (OTPs) offer a vast variety of products and services to travellers, such as flights, accommodations and ground transportation. Recently, offering a promotion such as a room upgrade, free meals, or discounted car rental has become a popular approach for offering customers more value for their money when completing a purchase. A promotion is expected to have a positive Average Treatment Effect (ATE) on outcomes like customer base growth or customer satisfaction, but can be viable only under set Return on Investment (ROI) constraints. 
ROI is a common measure in the field of marketing campaigns \cite{mishra2011current}, which represents the ratio of profit to investment as follows: 
\begin{equation}
ROI =  \frac{\Delta Revenue-\Delta Investment}{\Delta Investment}
\end{equation}
In order to ensure that a promotional campaign is sustainable over time, the target ROI is often restricted with a lower bound, such as $ROI\geq0$.
This creates a trade-off between the increment in target outcome and the cost of offering the promotion. 
A good example of such a promotion could be offering free lunch as an incentive for booking an accommodation; the promotional campaign will be sustainable only if the incremental value in accommodation booking rate can cover the extra cost of providing the free lunch. 
By offering the promotion to a subset of customers, a company can provide greater value to each customer while maintaining financial constraints that keep the promotional campaign sustainable over time \cite{reutterer2006dynamic}. 
One popular approach for efficiently assigning promotions to a subset of customers is targeted personalization \cite{lin2017monetary}.
In this approach, uplift models \cite{devriendt2018literature} are commonly used to estimate the expected causal effect of the treatment on the outcome in order to distinguish between the \textit{voluntary buyers} and the \textit{persuadables}, who would only purchase as a response to the incentive \cite{lai2006direct}. 
This can be achieved by estimating the Conditional Average Treatment Effect (CATE) of the promotion. $CATE_Y(x)$ is defined as the conditional increment in probability of completing the purchase $Pr(Y=1)$ \textit{caused} by the incentive, given the customer characteristics $x$.

CATE estimation can be conducted via various meta-learning techniques \cite{gutierrez2017causal}, such as the two-models approach \cite{radcliffe2007using}, transformed outcome \cite{jaskowski2012uplift}, uplift trees \cite{rzepakowski2012decision}, and other meta-learners \cite{kunzel2019metalearners}.
Some previous work aimed to address allocation of promotions by estimating the marginal effect \cite{zhao2019unified}, incorporating net value optimization within the meta-learner \cite{zhao2019uplift} and modeling the optimization as a min-flow problem \cite{makhijani2019lore}.
Such estimations might be unreliable \cite{lewis2015unfavorable,deng2015diluted}, however, due to high variance in the training dataset, which is heavily dependant on a majority of non-transactional visits having no intent to purchase. 
In addition, offline solutions might be biased towards historical data and require dynamic calibration \cite{zhou2008budget} according to long-term changes and seasonality trends, which are particularly common in the travel industry \cite{bernardi2019150} and in promotional campaigns \cite{kim2006pay}. 
Dynamic adaptive strategy adjustments is a common practice in cases of budget constraints and were found effective in ads management \cite{lin2017monetary}, influence maximization \cite{goldenberg2018timing, sela2018active}, marketing budget allocation \cite{badanidiyuru2012learning, lai2006direct, zhao2019unified} and discount personalization \cite{fischer2011dynamic}. 

In this paper we present a novel technique based on \textit{Retrospective Estimation}, a modeling approach that relies solely on data from positive outcome examples. 
This is extended by a method to dynamically maximize the target outcome of the treatment within ROI constraints. 
Our main contributions are:
(1) Optimization framework setup for uplift within ROI constraints;
(2) \textit{Retrospective Estimation} technique, relying only on data from positive examples;
(3) Dynamic model calibration for online applications;
and
(4) Large scale experimental validation.
The study proceeds as follows: The second section formalizes the problem we address and the constraints imposed by business and product requirements. 
The third section covers the suggested method for uplift modeling under ROI constraints, retrospective estimation technique, and dynamic calibration of the method. 
The fourth section covers validation of the method (Randomized Controlled Experiment), including details on data gathering, experimental setup and results.

\section{Problem formulation}
In this work, we focus on maximizing the overall number of customers completing a purchase, while deciding whether or not to offer the promotion in order to comply with the global constraint of $ROI\geq0$. In addition to $Y$ (a random variable representing the completion of a purchase), we consider the random variables $R$ and $C$, representing the monetary revenue associated with the purchase and the cost associated with the promotion, respectively. We assume that there are no revenue and no cost without a purchase.

We adopt the Potential Outcomes framework \cite{imbens2010rubin}, which allows us to express the causal effects of the promotions on all three different outcomes: for a customer $i$, $Y_i(1)$ represents the potential purchase if $i$ is given the promotion ($T=1$), while $R_i(1)$ and $C_i(1)$ represent the potential revenue and costs. Likewise, $Y_i(0)$, $R_i(0)$, and $C_i(0)$ represent the potential outcomes if the promotion is not given. We can thus define the conditional average treatment effect on all these variables, for a given customer with pre-promotion covariates $x$, as follows: $CATE_Y(x) = \mathbf{E}(Y_i(1)-Y_i(0)|X_i=x)$, $CATE_R(x) =\mathbf{E}(R_i(1)-R_i(0)|X_i=x)$ and $CATE_C(x) = \mathbf{E}(C_i(1)-C_i(0)|X_i=x)$.
We also define the profit outcome $\Pi_i=R_i-C_i$ and the loss as $\mathcal{L} = -\Pi$. Their associated $CATE_{\Pi}(x) = CATE_R(x)-CATE_C(x) = -CATE_{\mathcal{L}}(x)$ are simply the average increment in profit (or loss).
Our goal is to find a function $F$, that given $x$, decides if the customer is offered the promotion or not, while maximizing the total incremental number of transactions under incremental ROI constraints. Therefore, our problem is to learn $F$, given a sample of customers $U$, where some received the promotion and others did not. Formally, our objective is to learn $F$ from tuples $(x_i,T_i,Y_i,R_i,C_i)$,  $0<i<n$, where $T_i=1$ if the customer received the promotion and $T_i=0$ otherwise.

%there is a lot of whitespace
\subsection{Solution Framework}\label{sec:optimization}
We model $F$ with a scoring function $g$ and a threshold $\theta$, that outputs its decision as $ \mathbbm{1}{[g(x)\geq \theta}]$. 
Function $g$ needs to represent the increment in probability of booking as well as the associated revenue and costs.
By relying of existing data, we can reassign the promotions in $U$ by solving the following optimization problem:
\begin{equation}
\begin{array}{cc}
\noindent
\text{Maximize }  
\displaystyle \sum\limits_{i\in U} z_i \cdot CATE_Y(x_i)\\
\text{     subject to:    }
    \displaystyle
    \sum\limits_{i\in U} z_i \cdot CATE_{\mathcal{L}}(x_i)
    \leq 0

\end{array}
\end{equation}
Here, $z_i\in \{0,1\}$ is the assignment variable indicating if customer $i$ is offered the promotion or not. 
The target function is maximizing the total treatment effect, while the constraint specifies the condition of incremental total loss being non-positive (equivalent to positive ROI), by comparing the incremental costs and revenues potentially incurred by each incentivized customer. This sets a \textit{Binary Knapsack Problem} with negative utilities and weights.
We derive that $CATE_Y\leq 0 \implies CATE_{\mathcal{L}}\geq0$, since the revenue can only come from extra purchases.
Applying the transformation described in \cite{toth1990knapsack} results in the residual problem of a standard binary Knapsack Problem for customers with $CATE_Y > 0$ and $CATE_{\mathcal{L}} > 0$.
It has a new constraint constant given by $\sum_j CATE^j_{\mathcal{L}}$ for customers where $CATE^j_{Y}>0$ and $CATE^j_{\mathcal{L}}\leq0$.

According to the greedy algorithm for the \textit{Fractional Knapsack Problem}, we can approximate the solution by sorting customers $U$ by ${utility_i}/{weight_i}$ descending. In our problem this ratio maps to ${CATE^i_{Y}}/CATE^i_{\mathcal{L}}$. All customers are treated in that order, until the constraint is no longer satisfied. Denoting $j$ as the last treated customer, we can set $g(x)={CATE_Y(x)}/CATE_{\mathcal{L}}(x)$ and the threshold $\theta^*={CATE^j_{Y}}/{CATE^j_{\mathcal{L}}}$. Assuming the sample $U$ is a representative of future customers, we can use this rule to decide who to target with our promotion. This framework reduces the problem to learning $\frac{CATE_Y(x)}{CATE_{\mathcal{L}}(x)}$; $\sign[CATE_Y(x)]$; and $\sign[CATE_{\mathcal{L}}(x)]$. The following sections explore different approaches.

\section{Method for uplift modeling under ROI constraints}\label{sec:method}

We performed a comparison of two common uplift modeling techniques and introduced two novel modeling approaches:
\begin{itemize}
    \item \textbf{\textit{Two-Models}} \cite{radcliffe2007using}: a difference between two individual estimations of the effect under each of the treatments.
    \item \textbf{\textit{Transformed outcome}} \cite{jaskowski2012uplift}: direct estimation of $CATE_{Y}(x)$ with a single model.
    \item \textbf{\textit{Fractional Approximation}}: greedy sorting score relying on \textit{Two-Models} output - ${CATE_Y(x)}/{CATE_{\mathcal{L}}(x)}$.
    \item \textbf{\textit{Retrospective Estimation}} (\ref{sec:retro}): direct estimation of greedy sorting score, relying only on positive examples.
\end{itemize}
All methods rely on data from an online randomized controlled experiment \cite{kohavi2007practical, kaufman2017democratizing}, which allows us to write the potential outcomes $\mathbb{E}[Y_i(1)|X_i=x]$ as $\mathbb{E}[Y_i|T_i=1,X_i=x]$ (and their analogous for Revenue, Cost and Profit). These expectations are in turn estimated using machine learning. All methods are trained with underlying \textit{Xgboost} models \cite{chen2016xgboost} on data of >100 millions of interactions on Booking.com website, lasting over several weeks. The model performance was tuned with time-based cross validation and evaluated as described in section \ref{sec:eval}.
With ${\hat{C}(x)} =\mathbb{E}[C |x,T=1,Y=1] $ and ${\hat{R}_i(x)} =\mathbb{E}[R |x,T=i,Y=1] $, the full greedy criterion $\frac{CATE_Y(x)}{CATE_{\mathcal{L}}(x)}$ can be written in terms of expectations as follows:
\begin{equation}
\label{eq:greedy}
\frac{Pr(Y=1|x,T=1)-Pr(Y=1|x,T=0)}
{Pr(Y=1|x,T=1)[{\hat{C}(x)}-{\hat{R}_1(x)}]+Pr(Y=1|x,T=0){\hat{R}_0(x)}}
\end{equation}

\textbf{\textit{Fractional approximation}} technique relies on direct estimations of all the probabilities and expectations from formula \ref{eq:greedy}.  
The \textbf{\textit{Two-Models}} and \textbf{\textit{Transformed Outcome}} methods that are estimating only the numerator of Eq. \ref{eq:greedy}, are used as benchmark solutions to the unconstrained problem.
\subsection{Retrospective Estimation}\label{sec:retro}
This technique relies on the following response distribution factorization:
\begin{equation}
    Pr(Y=1|x, T=1) = \frac{Pr(T=1|x,Y=1)Pr(Y=1|x)}{Pr(T=1|x)}
\end{equation}
The denominator is the treatment propensity, which we can set to $0.5$. This decomposition does not allow us to estimate $CATE_Y(x)$ directly, but instead provides an expression for the ratio between $Pr(Y=1|x,T=1)$ and $Pr(Y=1|x,T=0)$:
\begin{equation}
     \frac{Pr(Y=1|x,T=1)}{Pr(Y=1|x,T=0)} = %\frac{2Pr(T=1|x,Y=1)Pr(Y=1|x)}{2Pr(T=0|x,Y=1)Pr(Y=1|x)}=
     \frac{Pr(T=1|x,Y=1)}{1-Pr(T=1|x,Y=1)} =
     \frac{S(x)}{1-S(x)}
\end{equation}
Here $S(x) = Pr(T=1|x,Y=1)$ is the probability of a positive example to result from the treatment $T=1$.
This allows us to write the full greedy criterion in terms of $S(x)$:
\begin{align}
&\frac{CATE_Y(x)}{CATE_{\mathcal{L}}(x)}= \frac{2S(x)-1}{{S(x) [{\hat{C}(x)}-{\hat{R}_1(x)}] + [1-S(x)] {\hat{R}_0(x)}}}\\
&\sign[CATE_Y(x)]=\sign\left[S(x)-0.5\right]\\
&\sign[CATE_{\mathcal{L}}(x)]= \sign\left[\frac{{\hat{R}_1(x)}}{2{\hat{R}_1(x)}-{\hat{C}(x)}}-S(x)\right]
\end{align}
Estimating $S(x)$, $\hat{R}_1(x)$,  $\hat{R}_0(x)$ and $\hat{C}(x)$ requires data only from customers where $Y=1$, making this approach robust to the noise in the general set of visitors (including scrapes and crawlers - a significant portion of traffic who rarely purchase) and less prone to class imbalance issues. It also requires a smaller transactional dataset, which is often easier to collect. 
Furthermore, assuming that the costs and revenues per booking are all independent of $x$ and that the revenue per booking is independent of treatment $T$ ($\hat{R}_i(x) = \hat{R}$ and $\hat{C}(x) = \hat{C}$) we can represent the ratio by equation \ref{eq:simplefrac}:
\begin{equation}
    \label{eq:simplefrac}
    \frac{2S(x)-1}{S(x) [\hat{C}-\hat{R}] + [1-S(x)] \hat{R}} 
\end{equation}
We can derive positive values of $CATE_{\mathcal{L}}(x)>0$ if $S(x)<\frac{\hat{R}}{2\hat{R}-\hat{C}}$ and $CATE_Y(x)>0$ if $S(x)>0.5$. 
This means that we can solve the greedy problem just by modeling $S(x)=Pr(T = 1|x, Y=1)$.

\subsection{Dynamic calibration strategy}
Fitting a function $g$ and a threshold $\theta$ offline provides a solution to the offline samples assignment.
However, during an online campaign, the general traffic distribution might differ and therefore the optimal threshold $\theta^*$ can shift accordingly.
We propose a simple curve fitting technique to address this problem. 
As seen in figure \ref{fig:roi}, we assume a non-linear monotonic relation between the portion of the exposed population $Q_{\theta}$ and the resulted $ROI$.
We suggest to model this relation as an exponential curve, by learning the parameters $a$,$b$ and $c$ such that: $\widehat{ROI}(Q)= ae^{bQ}+c$ using the Levenberg-Marquardt least-squares curve fitting algorithm \cite{more1978levenberg} to select a new $\theta^*$ such that $\widehat{ROI}(Q_{\theta^*}) = 0$.
Using this parametric curve fitting method allows incorporating historical and up-to-date data in ROI estimation, by refitting the curve every selected time-period.

\subsection{Evaluating an Uplift Model: Qini and Qini-ROI curves}\label{sec:eval}
\begin{figure*}%[H]
\centering
\begin{minipage}{0.48\textwidth}
  \centering
  \includegraphics[width=1\textwidth]{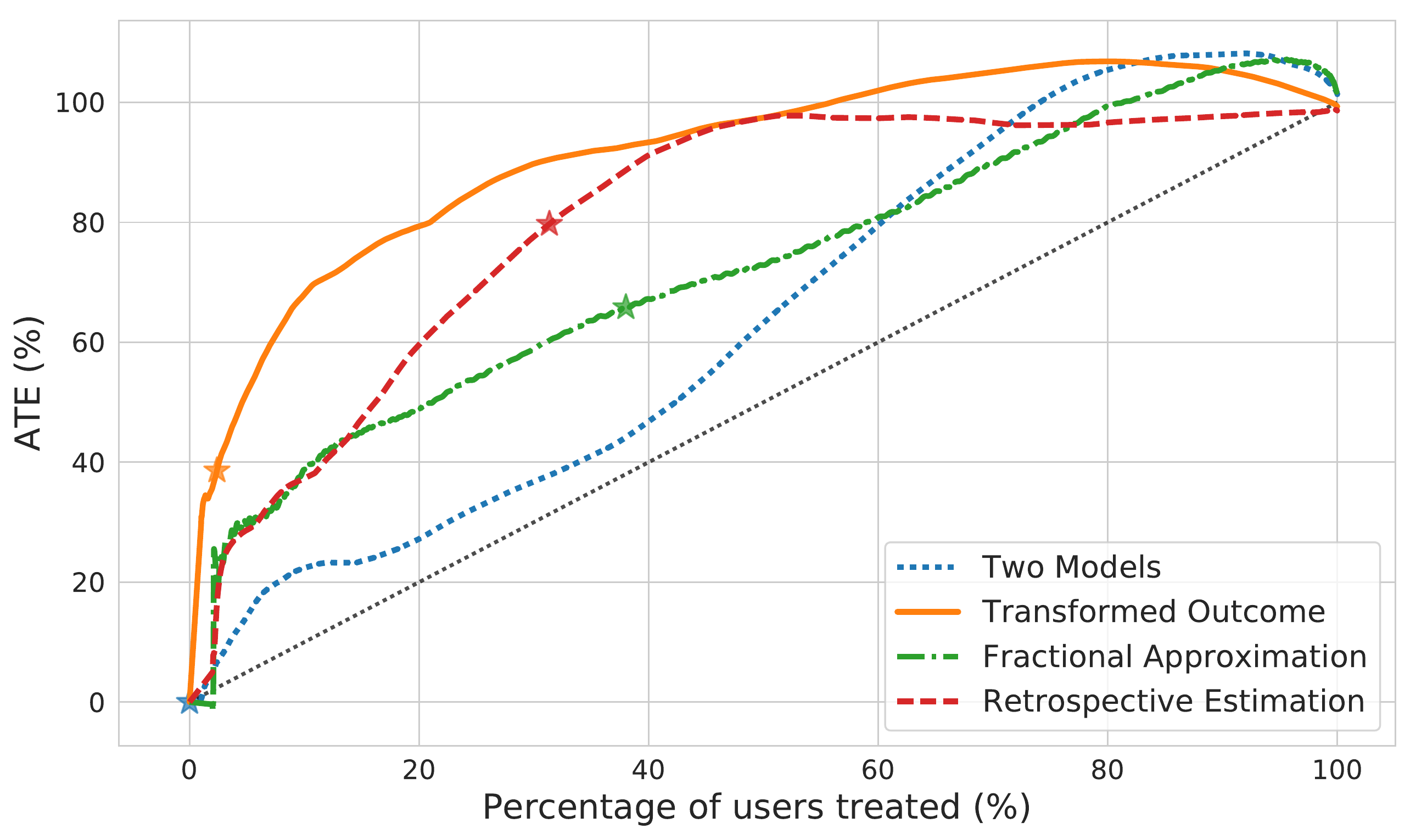}
  \captionof{figure}{Qini curves comparison}
  %\Description{Qini curves comparison}
  \label{fig:cate}
\end{minipage}%
\begin{minipage}{0.48\textwidth}
  \centering
  \includegraphics[width=1\textwidth]{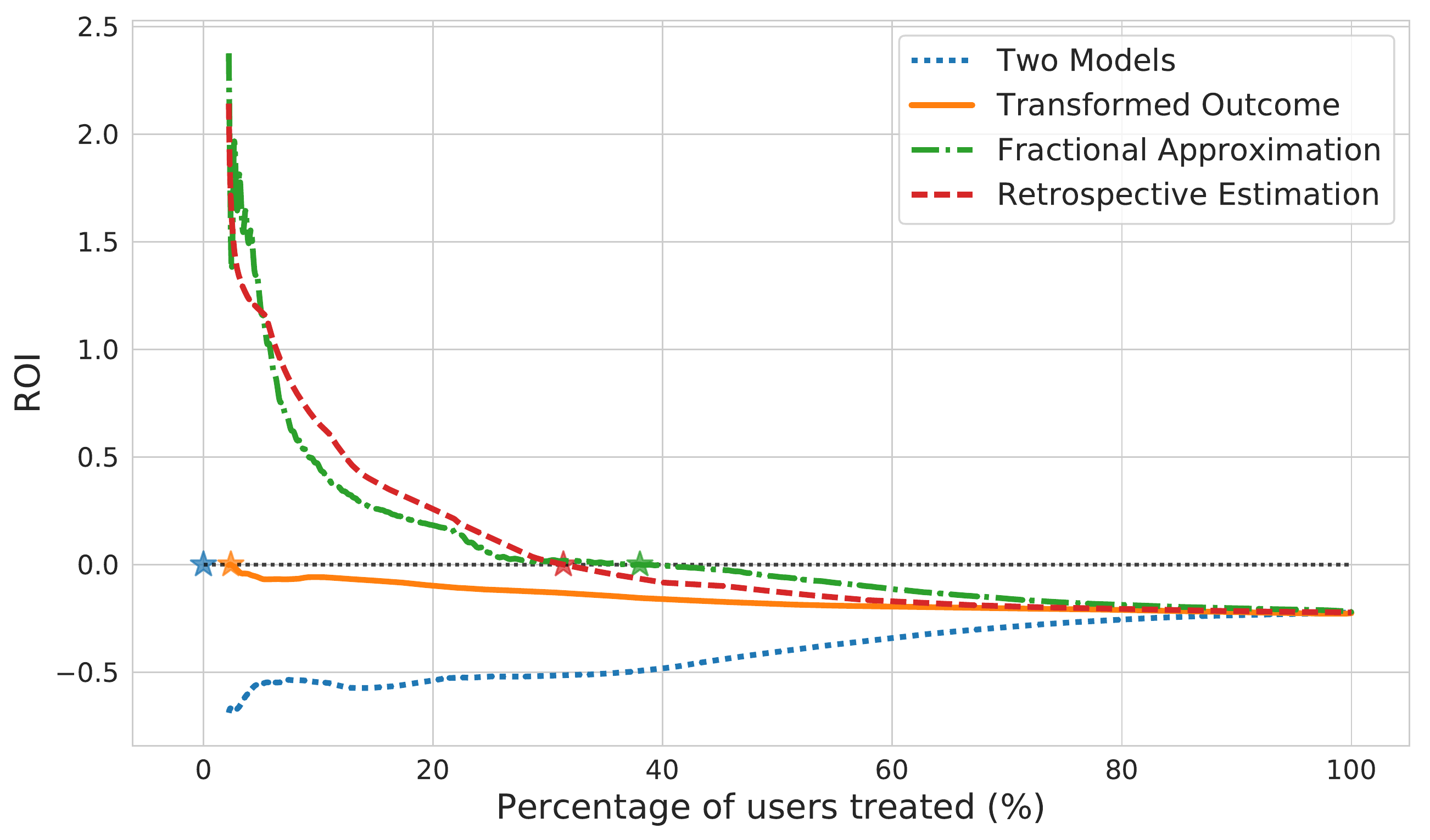}
  \captionof{figure}{Qini-ROI curves comparison}
  %\Description{Qini-ROI curves comparison}
  \label{fig:roi}
\end{minipage}
%\Description{ATE and ROI charts of different methods}
\end{figure*}

The methods are compared with \textit{Qini curves} \cite{radcliffe2007using} that represent the potential cumulative treatment effect achieved by targeting a selected top ranked population based on the model predictions (see figure \ref{fig:cate}).
The \textit{Qini Coefficient} (also called \textit{AUUC} - Area Under the Uplift Curve) \cite{devriendt2018literature} is used to quantify and evaluate the overall quality of an uplift model.

Following the same ranking procedure, we propose the \textit{Qini-ROI curve}, representing the expected ROI that can be achieved by targeting a selected portion of the top ranked population (see figure \ref{fig:roi}).
According to the described optimization problem (\ref{sec:optimization}), $ROI\geq0$ is a binding constraint to the treatment assignment, therefore the optimal model threshold is at the operating point where the treatment effect is the highest and also $ROI\geq0$.

Thus, the models' performance is evaluated by \textit{Maximal $ATE$ at $ROI\geq0$} metric. The \textit{AUUC} and \textit{Maximal targeted population at $ROI\geq0$} metrics are used to evaluate the class-separation and the population coverage of the methods.

\section{Experimental Study}

% This section describes the experimental setup, the underlying data, and the results of the following experiments:
% (1) Randomized controlled trial of undiscriminated treatment assignment;
% (2) Uplift methods training and offline evaluation;
% (3) Randomized controlled trial of personalized treatment assignment.

\subsection{Randomized controlled trial of undiscriminated treatment assignment} \label{sec:randomized}
In order to test the suggested method, we conducted a series of offline and online experiments. As a preceding step to our method, we conducted a randomized controlled trial to assess the potential impact of a given promotion.  
The experiment took the form of an online A/B test, in which the treatment group was offered a promotion and the control group was not. The experiment was conducted on Booking.com website and lasted for several weeks.
In our experiment, all of the customers in the treatment group received an offer of a free benefit as an incentive for booking an accommodation. The individual outcome metrics - completing a booking, promotion cost, and incremental revenue - were aggregated and compared between the control (no promotion) and treatment (free benefit) groups, producing an estimation of the $ATE$ and $ROI$ of the promotion treatment.
The experiment resulted in a conclusively positive treatment effect \textbf{($ATE>0$)}, but with an \textbf{$ROI<0$} value, which is insufficient in order to keep offering the promotion.

\subsection{Uplift methods training and offline evaluation}

The data gathered during the A/B test was used as training instances for uplift modeling.
The dataset consisted of >100 million website visit instances, associated with customer's current and historical search queries and transactions.
Each instance was attributed with associated transaction, revenue, and cost, if applicable.
We evaluated the models' performance on a validation set according to the metrics described in \ref{sec:eval}.
Figure \ref{fig:cate} presents the comparison of the models' uplift performance; Figure \ref{fig:roi} presents the comparison of the models' ROI profile; key values are summarized in table \ref{tab:results}. 
The stars at each curve represent the best operating points (in terms of $ATE$) for each method, s.t. $ROI\geq0$.
% ROI and QINI figures

\begin{table}[b]%[H]
  \caption{Uplift methods' performance comparison}
  \label{tab:results}
  \begin{tabular}{lccc} %c
    %\toprule
    \hline
    Method &
    AUUC &
     \specialcell{Max. Population\\at ROI=0} &
     \specialcell{Max. $ATE$ \\at ROI=0} \\
     \hline
    %\midrule
    Two-Models               & 0.638 &  0\%     & 0\%\\ %& 108.2\% \\
    Transformed Outcome      & \textbf{0.912} & 2.4\%   & 38.6\% \\ %& 107.2\$ \\
    Fractional Approximation & 0.721 & \textbf{38\%}     & 65.8\%\\ %& 106.9\%  \\
    Retrospective Estimation & 0.806 & 31.4\%   & \textbf{79.7\%}\\ %& 98.8\% \\
    \hline
  %\bottomrule
\end{tabular}
\end{table}

 The \textit{Transformed Outcome} method achieves the highest $AUUC$ score (0.912), however, its $ROI=0$ intersection is only at 2.4\% of population, achieving 38.6\% of the treatment effect relative to giving the promotion to all the users.
The \textit{Two-Models} method has the highest potential treatment effect (109\%) in an unconstrained setup, but it performs poorly in key metrics and doesn't have any operating point with $ROI\geq0$.
The \textit{Fractional Approximation} method provides the highest population coverage (38\%) at $ROI\geq0$ and achieves 65.8\% of the full treatment effect.
Recalling the objective function of the optimisation problem, the \textit{Retrospective Estimation} method provides the best solution.
It leads with a great advantage on maximal treatment effect achieving 79.7\% of the possible uplift and is a runner-up in both AUUC (0.805) and Population at $ROI=0$ (31.4\%) metrics.
It is worth highlighting that the \textit{Retrospective Estimation} performance relies solely on positive examples data.

\subsection{Randomized controlled trial with personalized treatment assignment}

We recreated the randomized controlled trial setup as an online validation of our technique \ref{sec:randomized}.
For this iteration, visitor traffic was split into four randomly-assigned treatment groups:
\begin{itemize}
    \item \textbf{A} - \textbf{\textit{Control Group}} - no promotion offering
    \item \textbf{B} - \textbf{\textit{Undiscriminated Treatment}} - promotion is offered to everyone
    \item \textbf{C} - \textbf{\textit{Personalized Treatment}} - promotion is offered to a population with $Score(x)\geq\theta$
    \item \textbf{D} - \textbf{\textit{Dynamically Personalized Treatment}} - promotion is offered to a population with $Score(x)\geq\theta_{d}$
\end{itemize}
We used the \textit{Retrospective Estimation} technique for the underlying personalized treatment assignment method in groups \textbf{C} and \textbf{D}, due to its superior performance on the target metric.
Model score threshold ($\theta$) was selected based on offline evaluation of the model, such that $31.4\%$ of the population are expected to receive the promotion.
We re-calibrated the threshold for the \textit{Dynamically Personalized Treatment} ($\theta_d$) after each predefined time-period.
A randomized controlled trial (over 8 update periods) was conducted under the same promotion conditions as in the first experiment (an offer of a free benefit as an incentive for an accommodation booking on the website).
The goal of the experiment was to evaluate the incremental ATE and the ROI of the treatments groups compared to  control group \textbf{A}.
\begin{figure*}%[b]
  \centering
  \includegraphics[width=0.95\textwidth]{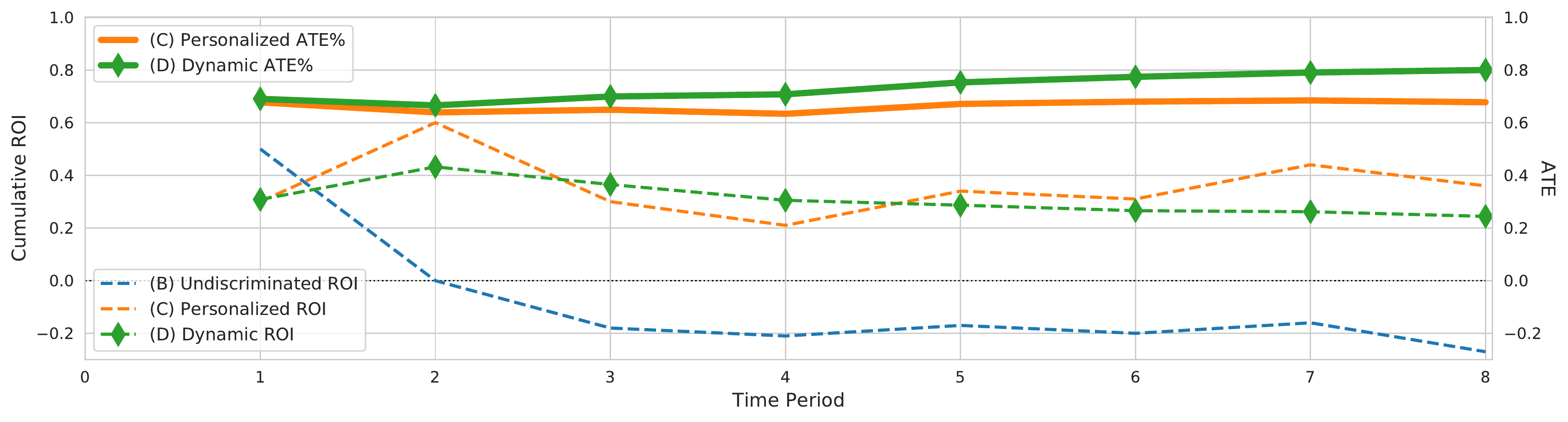}
  \caption{Experimental results (achieved ROI and portion of ATE) of personalized promotions assignment}
  %\description{Experimental results of personalized promotions assignment}
 \label{fig:dynamic}
\end{figure*}
Figure \ref{fig:dynamic} presents the measured cumulative $ROI$ (dashed lines) and the relative $ATE$ (solid lines) of the treatments, where the \textit{diamond} markers ($\Diamond$) represent the dynamic calibration points during the experiment.
Treatment \textbf{B} resulted in $ROI$ values below 0, similarly to the previous iteration of the experiment.
Treatment \textbf{C} (Personalized group) provided consistently high $ROI$ values ($ROI=0.36$ at the last time period) and its total cumulative uplift is 66\% of that achieved by fully-treated group \textbf{B}.
Treatment \textbf{D} starts with a high $ROI=0.3$ in the first period and converges to $ROI=0.24$ at the end of the experiment.
Dynamic adjustment of the threshold in group \textbf{D} allowed for exposing the treatment to 47\% of the population, achieving on average 74\% of the uplift achieved by treatment \textbf{B}, including 80\% during the last period.

\section{Conclusions}

In this work, we introduced an optimization framework to solve the personalized promotion assignment by relying on the Knapsack Problem and presented a novel method to estimate the required sorting score.
Classical uplift models excelled in AUUC metric, but under-performed in ROI-related measures as they do not account for the cost factor.

Our novel \textit{Retrospective Estimation} technique relies solely on data from positive outcome examples, which makes its training easier and more scalable in big-data environments, especially for cases where only the outcomes' data is available.
The method achieved 80\% of the possible uplift, by targeting only 31\% of the population and meeting the $ROI\geq0$ constraint during offline evaluation.
It outperforms the direct \textit{Fractional Estimation} model, potentially thanks to robustness to noise in the training data, which opens an opportunity for further research on new applications.

We suggested a dynamic calibration technique, which allows adjusting the model threshold and exposing a different portion of the population to the treatment, by relying on online performance.
We evaluated our suggested techniques through a massive online randomized controlled trial and showed that a personalized treatment assignment can turn an under-performing promotions campaign into a campaign with a viable $ROI$ and significant uplift ($ATE=66\%$ of the possible effect).
Applying the dynamic threshold calibration to online data improved the uplift of the treatment by up to 20\%, while staying within the constraints, and provided a long term solution for a rapidly changing environment.

Our contributions lay the foundation for future research to apply the \textit{Retrospective Estimation} technique to other uplift modeling and personalization applications. This sets the groundwork for new optimization approaches and provides solid empirical evidence of the benefits of online personalized promotion allocation and dynamic calibration.
It creates more value to the customers and increases the loyal customer base for the vendors and the platform.

\begin{acks}
We would like to thank the entire Value Team for their contribution to the project and to Adam Horowitz, Tamir Lahav, Guy Tsype, Pavel Levin, Irene Teinemaa and Nitzan Mekel-Bobrov for their support and insightful discussions.
\end{acks}

\bibliographystyle{ACM-Reference-Format}
\bibliography{camera_ready_sub}

\end{document}